%% file: wavclumps.tex
\documentclass[10pt]{article}

\input{preamble.tex}

\usepackage{lmodern}
\usepackage[utf8]{inputenc}
\usepackage{graphicx}
\usepackage{times}
\usepackage{amsmath}
\usepackage{indentfirst}
\usepackage{algorithmicx}
\usepackage{algorithm}
\usepackage[noend]{algpseudocode}
\usepackage{setspace}

\begin{document}
\noindent

\title{Multiresolution and Hierarchical Analysis of Astronomical Spectroscopic Cubes using 3D Discrete Wavelet Transform}

\authorname{Martín Villanueva* and Mauricio Araya*}
\authoraddr{*Universidad Técnica Federico Santa María, Valparaíso, Chile}


\maketitle

\keywords
Source Identification, Multiresolution Analysis, Wavelets, Hierarchical Relations.

\abstract

The intrinsically hierarchical and blended structure of interstellar molecular clouds, plus the always increasing resolution of astronomical instruments, demand advanced and automated pattern recognition techniques for identifying and connecting source components in spectroscopic cubes. We extend the work done in multiresolution analysis using Wavelets for astronomical 2D images to 3D spectroscopic cubes, combining the results with the Dendrograms approach to offer a hierarchical representation of connections between sources at different scale levels. We test our approach in real data from the ALMA observatory, exploring different Wavelet families and assessing the main parameter for source identification (i.e., RMS) at each level. Our approach shows that is feasible to perform multiresolution analysis for the spatial and frequency domains simultaneously rather than analyzing each spectral channel independently.

\section{Introduction}

Astronomy is an important source of challenges for pattern recognition, basically because the objective of an astronomical observation is always to detect and quantify patterns. Source extraction \cite{Bertin} is the first step for most of the astronomical applications such as transient planet detection, galaxy formation, stellar evolution, disks/jets dynamics or molecular cloud analysis. This last one consist in detecting more or less diffuse but overlapped structures over ``cloudy'' regions of the space composed by cold dust and gas \cite{Williams}. These fuzzy structures are called clumps, which are the precursors of core structures that will become more classical astronomical sources such as disks, stars or planets, allowing to study the very early formation of them. A key information that these cold structures provide is their molecular composition, because emission lines are sparsely distributed in the low-energy electromagnetic spectrum. In other words, each molecule emits in different frequencies that are easier to disentangle compared to studying hotter phenomena. Therefore, the instruments that study these objects (such as radiotelescopes) usually provide high-resolution spectral information, which produce large three-dimensional data cubes (i.e., spectroscopic cubes) where these structures need to be detected both within the spatial and spectral axes.  

Several specific algorithms have been proposed for this goal, from those that are strongly based on the underlying astrophysical process \cite{Stutzki}, to those that directly exploits the pixel-based structure of the observations \cite{Berry}. In recent years, the intrinsic hierarchical structure of molecular clouds has been exploited through the use of hierarchical algorithms and data structures such as dendrograms \cite{Rosolowsky}. 
On the other hand, multiresolution analysis of images has been studied thoroughly for analyzing 2D images in astronomy \cite{Starck}, including the use of Wavelets functions.

In this paper we extend the work done in multiresolution analysis using Wavelets for determining the hierarchical relations between clumps on both spatial and frequency domains at different levels of detail.
The main novelty is performing wavelet decomposition, reconstruction and visualization directly over 3D data cubes using a less memory-consuming approach.

In Section \ref{definition} we describe the addressed problem and present a non-exhaustive state-of-the-art discussion for clump identification and their hierarchical relations. Section \ref{proposal} introduces our approach step by step, while Section \ref{experiments} presents the experiments and analysis of our results.
At last, Section \ref{conclusion} presents the conclusions and future work.

\section{The Clump Detection Problem} \label{definition}

We address the problem of automatically identifying clump structures given a 3D spectroscopic cube, while determining the hierarchical relationships between them. 
A clump can be generally understood as a region of high density within a larger area of lower density, in relation to the star formation in molecular clouds \cite{Kennicutt}. These molecular clouds are usually non homogeneous in their density, and are composed of cores of high density gas matter contained and separated by low density areas. Each observation of such clouds, may contain multiple source cores and clumps in complex configurations both in spatial and spectral domains.

Manual identification of clumps becomes impractical for 3D cubes with high-resolution axes, and as the number of observations increases with each new instrument, the automation of this process becomes critical.
Moreover, as the image becomes more crowded with clumps, problems like blended emissions arise, where multiple dense cores converge in a nearby area. In these cases, subjective biases become more evident as each astronomer might perceive the data differently. Therefore, astronomers rely on automated clump detection algorithms, to remove individual biases and provide consistent and objective results.

%

\begin{figure}[htpb!]
\centering
\includegraphics[width=.99\linewidth]{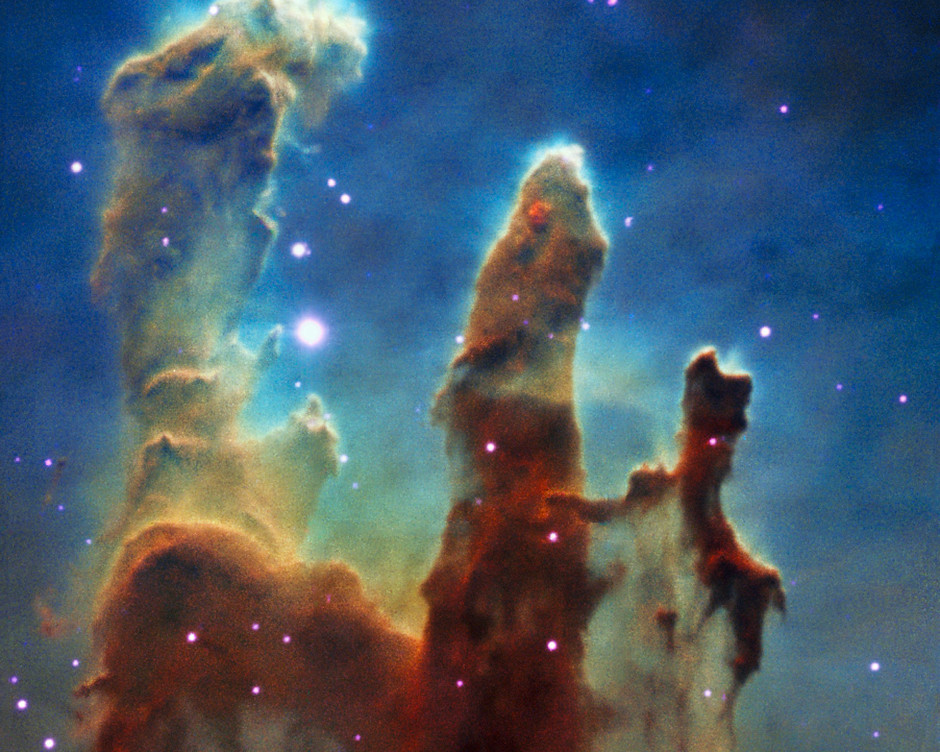}
\caption{Pillars of Creation. Interstellar cloud of gas and dust in the Eagle Nebula, known for its complexity and beauty. Original image by ESA/Hubble Space Telescope.}
\label{fig:pillars}
\end{figure}



But identifying clumps is not enough: each dense clump is often contained on lower density cores, constituting a hierarchical structure (see Figure \ref{fig:pillars}). Consequently, the challenge also includes determining the hierarchical relationships between clumps, because this information is crucial to understand processes such as \textit{star formation} \cite{Alves}. In the rest of this section we discuss first those clumping algorithms that do not describe these relationships, and then the ones that they do.






\subsection{Classical Clumping} \label{clumping}
\label{clumping}    
    
The \emph{GaussClumps} algorithm \cite{Stutzki} was the first successful approach to handle the problem of automatically detecting clumps. The main idea was to adjust Gaussian profiles that best fits each of the emission peaks. 
The algorithm iteratively searches for the current emission peak, and tries to fit a Gaussian function 
through an error minimization process. Then, the Gaussian is subtracted from the data cube, and the same step is repeated over the residual cube. The result is a set a Gaussians with different positions, orientations and intensities. The sum of all this Gaussian components plus the background noise, allows us to reconstruct the data cube. In this algorithm each Gaussian is considered as a single clump, and because Gaussians could overlap between them it is possible that pixels are assigned to more than one clump.


The \emph{ClumpFind} algorithm \cite{Williams} simulates how the human eye decomposes the maps into clumps by looking at the contour levels in a gradually and decreasing way. The algorithm works by contouring the data at multiples of the Root Mean Square (RMS) noise of the observations and isolating structures at each level. Then, it connects the structures between different levels through their common pixels. 

The result is a \textit{Clump Assignment Array} (CAA) structure, where each pixel is assigned exclusively to one clump.
The key feature of ClumpFind is that no a priori clump profile is assumed, and then the identified clumps can have any shape.


The more recent \emph{FellWalker} algorithm \cite{Berry}, uses a gradient-tracing scheme similar to \textit{hill-climbing} to reach a local emission peak, which defines a new clump. FellWalker, like ClumpFind, segments the supplied data cube into a number of disjoint regions, each associated with a single significant peak. The algorithms works by computing ascent paths for each single pixel. These paths may reach a local peak and therefore finding a new clump, or
reach a previous ascent path, which results in assigning the pixels on the current path to the corresponding (crashed) clump. At last, the detected clumps with low dip between them are merged as a single clump. The resulting structure is the same of ClumpFind (i.e., CAA).


\subsection{Hierarchical Relations}

	The \emph{DendroFind} approach \cite{Rosolowsky} tracks the hierarchical structure by computing \textit{isosurfaces} (surfaces with the same intensity) like ClumpFind, but indexing their relationships in a tree data-structure also know as Dendrograms.
	
    The elements in the Dendrogram structure correspond to specific volumes in data cubes, defined by their bounding isosurfaces.
    The order and structure of the tree, defines the physical relations between the identified cores.

    \emph{Multiresolution Analysis} (MRA) it is a technique used to analyze astronomical data images at different levels of detail \cite{Starck}. The Wavelet Transform, which maps input data to the \textit{scale-position} Wavelet space, is one of the most used approaches nowadays. In particular, \textit{Discrete Wavelet Transform} (DWT) can be used to transform images with no loss, allowing to process different levels of details independently.
    A special type of DWT is used in the practice, namely the \textit{Stationary Wavelet Transform} (SWT), also called \textit{À trous} wavelet transform. It consist on a redundant version of DWT, that has the property of mapping input images to images in the wavelet space with the same size (because of the redundancy). 
     This technique has been used for the study of the stellar \textit{Initial Mass Function} (IMF) in Alves et al. \cite{Alves}, by applying the SWT to images of interstellar molecular clouds.
     The clump identification procedure first maps the corresponding (2D) image, to the wavelet space at different scales. For a given scale $i$, structures are isolated at $3\sigma_i$, with $\sigma_i$ being the noise of the image at scale $i$. Then a detected structure at scale $i$ is connected with a structure at scale $i+1$ if its local maximum drops in the structure at scale $i+1$.

    Gregorio el al. \cite{Gregorio} follows a very similar approach, performing a SWT over 2D images to get a multiresolution view of the data. However they were not satisfied with the thresholding step of above, so they use clump identification algorithms (like ClumpFind) to segment the data at the different levels on the Wavelet space. 

\section{Wavelet-based Approach to 3D Clumping} \label{proposal}

\begin{figure}[htpb!]
\centering
\includegraphics[width=9cm]{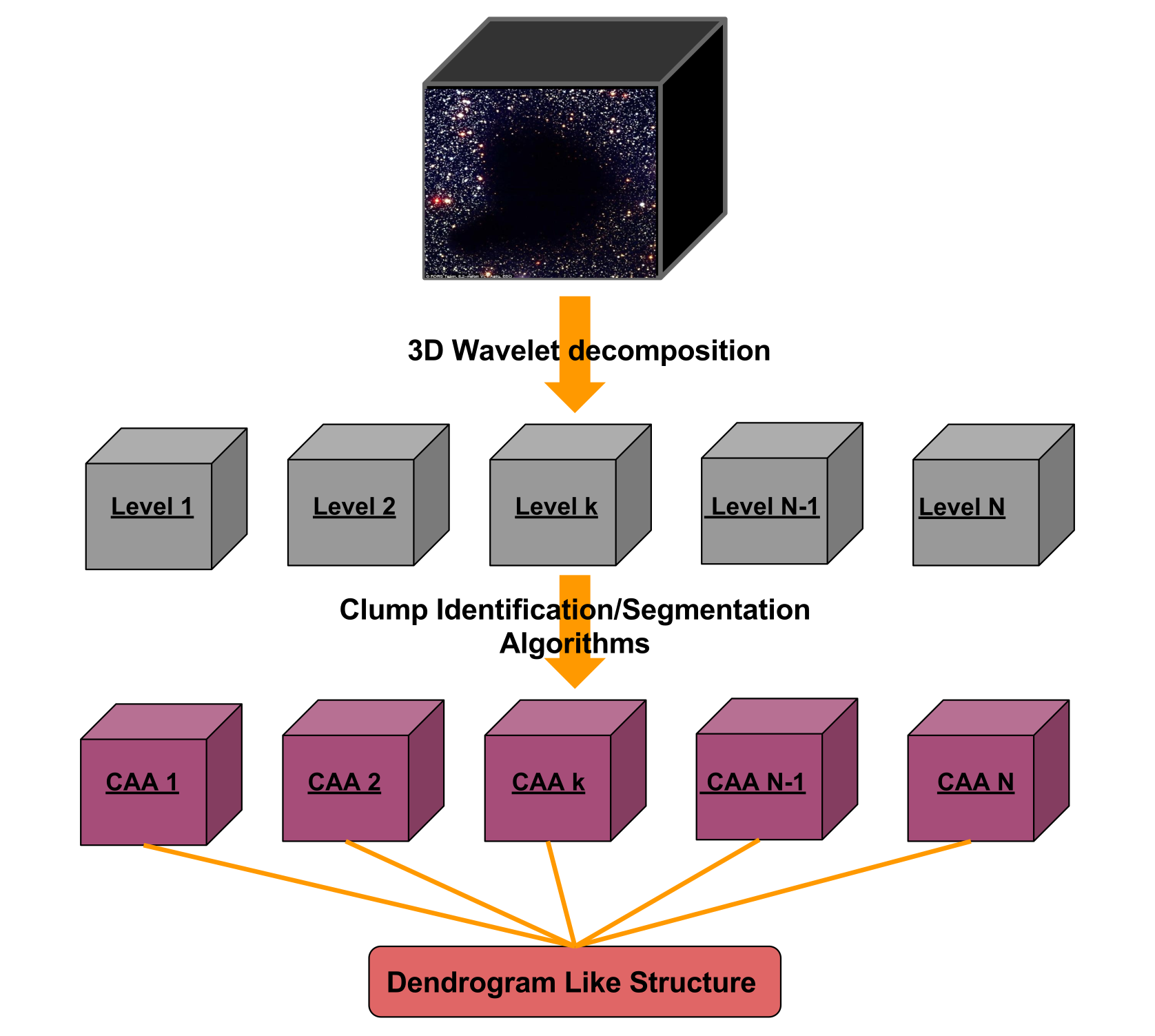}
\caption{Proposed Solution Outline}
\label{fig:wavclump}
\end{figure}

We propose extending the ideas of Alves et al. \cite{Alves} and Gregorio et al. \cite{Gregorio} to support 3D data directly. The core idea is to perform MRA by applying the 3D version of the original Discrete Wavelet Transform (3D-DWT) at different scales (in space and frequency) rather than using the SWT, and then reconstructing the cubes with different levels of details. After computing the low level detail cubes with the 3D-DWT, a clump identification/segmentation algorithm is applied to each of these cubes, indexing the sources found at different levels into a tree-like Dendrogram structure. An outline of this procedure is shown on Figure \ref{fig:wavclump}.

\subsection{DWT and Multiresolution Analysis.}

The DWT of a one-dimensional signal $x(t)$ is computed by convolving it through a series of low and high pass filters. If we denote $g$ the impulse response of a \textit{low pass filter} and $h$ the impulse response of a \textit{high pass filter}, then we compute the DWT coefficients as follows: 
\begin{align}
\label{eq:1dDWT}
\begin{split}
    y_{\textbf{low}}[n] &= (x*g)[n]=\sum \limits _{{k=-\infty }}^{\infty }{x[k]g[n-k]} \\
    y_{\textbf{high}}[n] &= (x*h)[n]=\sum \limits _{{k=-\infty }}^{\infty }{x[k]h[n-k]}
\end{split},
\end{align}
where we call $y_{\textbf{low}}$ the approximation coefficients, and $y_{\textbf{high}}$ the detail coefficients. Since in each case half the frequencies have been dropped, then half the samples in $y_{\textbf{low}}$ and $y_{\textbf{high}}$ can be removed according to Nyquist-Shannon sampling theorem. This process can be repeated taking  $y_{\textbf{low}}$ as the signal, and iteratively splitting it into approximation and detail coefficients.

\begin{figure}[htpb!]
\centering
\includegraphics[width=9cm]{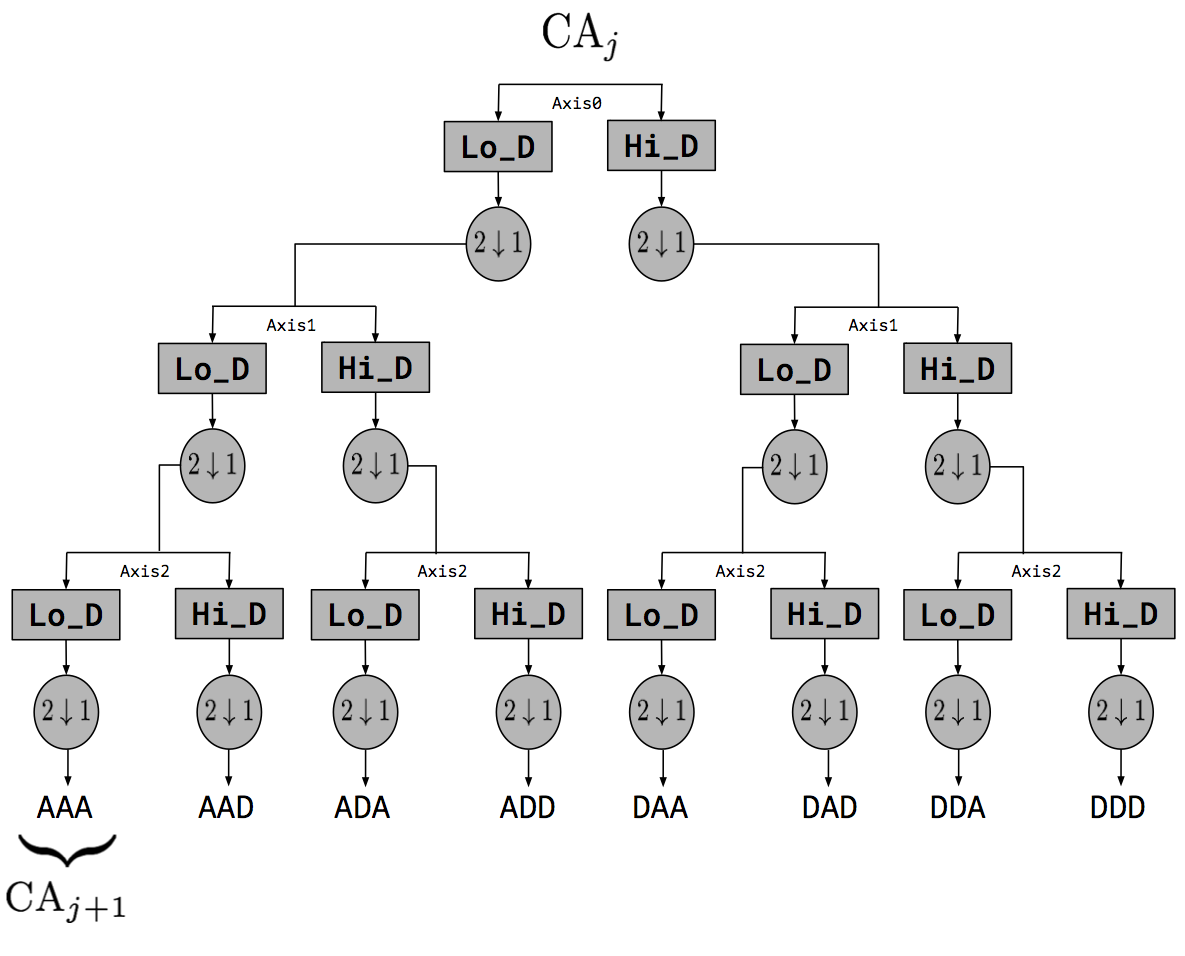}
\caption{3D Discrete Wavelet Transform scheme}
\label{fig:3Ddwt}
\end{figure}

In the case where the signal is a 3D data cube, an extension for the procedure described in \eqref{eq:1dDWT} exist. As shown in figure \ref{fig:3Ddwt}, each axis of the cube can be treated separately, i.e., we apply the \textit{Low Decomposition Filter} (\texttt{Lo\_D}) and \textit{High Decomposition Filter} (\texttt{Hi\_D}) plus the down sampling ($2 \downarrow 1$) through the \textit{Axis0} first, and then repeat this procedure for \textit{Axis1} and \textit{Axis2}, giving us as results eight sub-cubes of coefficients: \{AAA, AAD, ADA, ADD, DAA, DAD, DDA, DDD\}. Here $\text{A}$ stands for low-pass approximation coefficients, and $\text{D}$ stands for high-pass detail coefficients. Since we want to remove high frequency details from the data, we drop all these coefficients but $\text{CA}_{j+1} = \text{AAA}$, which stores low frequency details only. 

Please note that the starting point $\text{CA}_0$ is equal to the original data cube, but at each level of decomposition the coefficient cubes are reduced in $1/8$ of the size, making this approach memory efficient. Also, the computation of all the detail coefficients can be completely avoided, by just performing the operations in the most left branch of the tree of Figure \ref{fig:3Ddwt}. This idea follows from the fact the 3D-SWT is not an option, because the coefficient cubes would have same size of the original cube (due to its redundancy). For big enough data cubes it will not be able to compute more than couple of decomposition levels because of memory consumption.


\begin{figure}[htpb!]
\centering
\includegraphics[width=9cm]{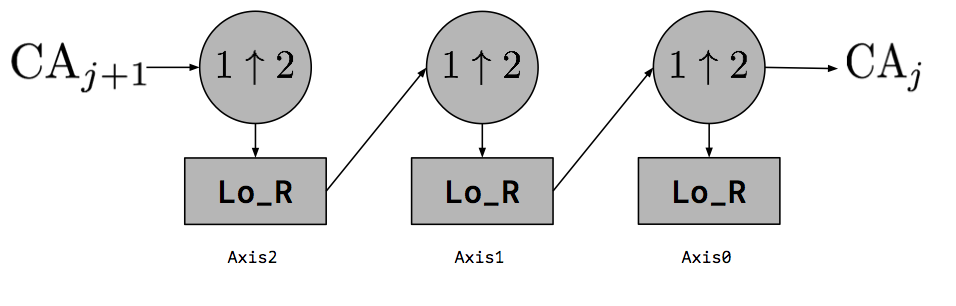}
\caption{3D Inverse Discrete Wavelet Transform scheme}
\label{fig:3Didwt}
\end{figure}

However, since we need to produce images of the same size of the original data cube, the approximation coefficients $\text{CA}_j$ cannot be used directly, because these are reduced representations on the Wavelet space. What can be done is to compute the Inverse Discrete Wavelet Transform (IDWT) for each level, and using them as the multiresolution representation. This inverse transform consist in applying the inverse process from \textit{Axis2} to \textit{Axis0}, i.e., the values on the corresponding axis are first upsampled by $2$ (filling with $0$'s) and then a \textit{Low Pass Reconstruction} filer (Lo\_R) is applied through this axis, as shown in Figure \ref{fig:3Didwt}. 

In order to get coherent results, an appropriate 3D Wavelet family has to be chosen \cite{Daubechies}, such that the features computed at the different levels are meaningful. These are empirically studied in section \ref{experiments}.

\subsection{Sources Identification} \label{SourceIdentification}

In general, any source identification, segmentation or clumping algorithm could be applied: from thresholding, watershed algorithms, or other advanced computer vision techniques to the clumping algorithms described in section \ref{clumping}. Moreover, it is not strictly necessary to use the same algorithm over all the cubes (data and approximations), but it is possible to apply the algorithm that performs better in a given level representation. As low level representation in the Wavelet space have fine-grained details, they might need robust (and computationally expensive) algorithms to perform a proper source identification step. On the other hand, images in a high level representation have coarse-grained details, so simple (and computationally cheaper) algorithms such as thresholding segmentation might be enough. 


In this paper we use the same clumping algorithm (i.e., FellWalker) over all the level representations, yet we use a different RMS parameter at each level. Indeed, \textit{all the clumping algorithms are very dependant of the RMS value of the input cube}, so this idea can be applied to any clumping algorithm. This is needed because we are dropping out all the high-pass coefficients for the reconstruction process, so the high frequency contributions to the signal are lost. As the RMS is a measure of the mean intensity of the signal, then the RMS will be a decreasing function over the level of decomposition.

\subsection{Hierarchical Representation}

The computed CAAs for the cubes at each decomposition level can be used to connect common structures found between neighbor levels. In \cite{Alves}, they verify if the peak pixel of a clump detected at level $i+1$ is contained in a clump detected at level $i$, forming the hierarchical relation between them.
However, for clumps with noisy peaks or clumps that are very flat, the peak pixel is not a good clump centrality indicator. Therefore, we propose to use the \textit{centroid} of a clump, computed as:
    \begin{equation*}
        \textit{centroid} = \sum_{i \in \text{clumpIndexes}} \text{pixel}_\text{position}[i] \cdot \text{pixel}_\text{intensity}[i],
    \end{equation*}
    where $\text{clumpIndexes}$ are the indexes of the pixels that belong to the corresponding clump. 

Lets call $\text{L}\{i\}\text{C}\{p\}$ to the $p$-th clump detected at decomposition level $i$, and $\text{L}\{i+1\}\text{C}\{q\}$ to the $q$-th clump detected at decomposition level $i+1$. If they were hierarchical related by one of the criteria presented above, we will represent this relation in a tree structure like the one shown in figure \ref{fig:3Dplot} (left).

\section{Experiments and Results} \label{experiments}

For the following experiments we use data from the ALMA  science verification archive\footnote{https://almascience.nrao.edu/alma-data/science-verification}. 
Specifically, we use an observation of a methanol emission line of the Orion KL region, which consists of a cube with $100 \times 100$ pixels for spatial resolution, and $41$ frequency channels.

For the Wavelet MRA we have chosen six different wavelet families in order to see the differences in the obtained low-frequency approximations: \textit{Haar}, \textit{Daubechies}, \textit{Symlet}, \textit{Coiflet}, \textit{Biorthogonal} and \textit{Reverse Biorthogonal}. In each case, the number of vanishing moments  was set according to the MATLAB Wavelet Toolbox indications\footnote{www.mathworks.com/help/wavelet/ug/wavelet-families-additional-discussion.html},
plus experimental tuning.

To compute the clumps at each level of decomposition, we use the Fellwalker algorithm \cite{Berry} implemented in ACALIB\footnote{https://github.com/ChileanVirtualObservatory/ACALIB}.


\subsection{Thresholding at Different Levels}



\begin{figure}[htpb]
    \centering
   \includegraphics[width=.99\linewidth]{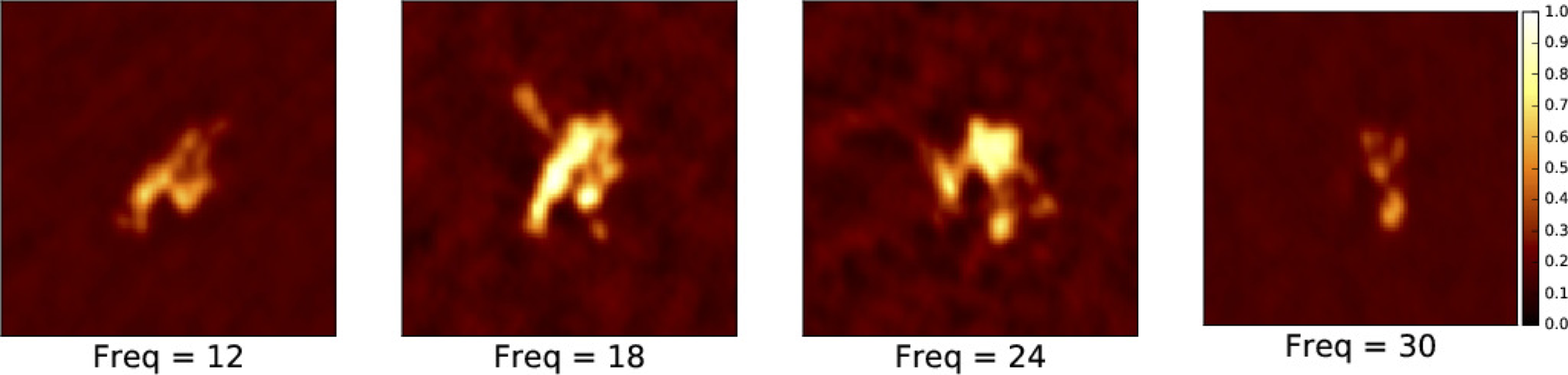}\\
   (Original)\\
      \includegraphics[width=.99\linewidth]{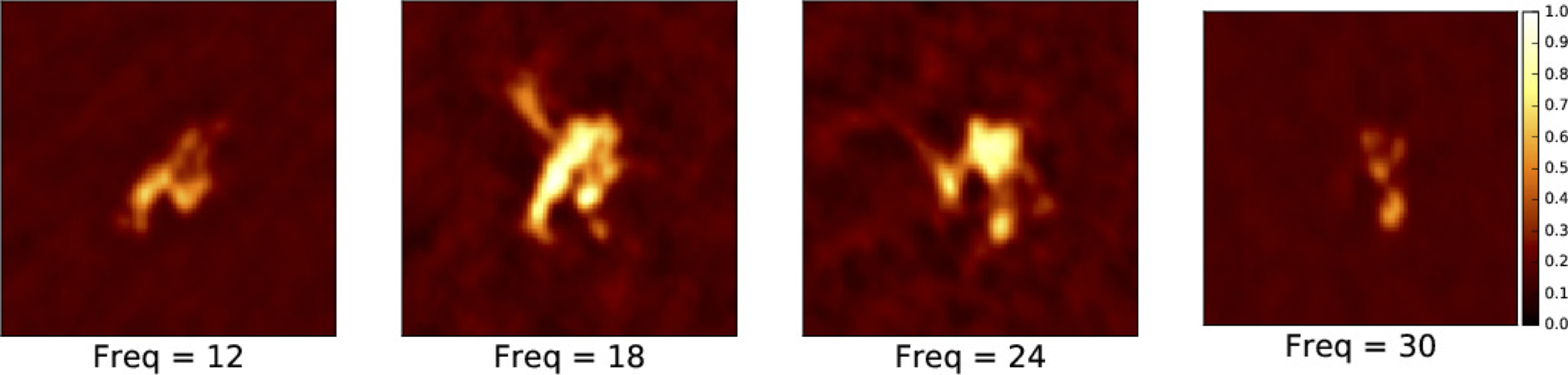}\\
   (Level 1)\\
   \includegraphics[width=.99\linewidth]{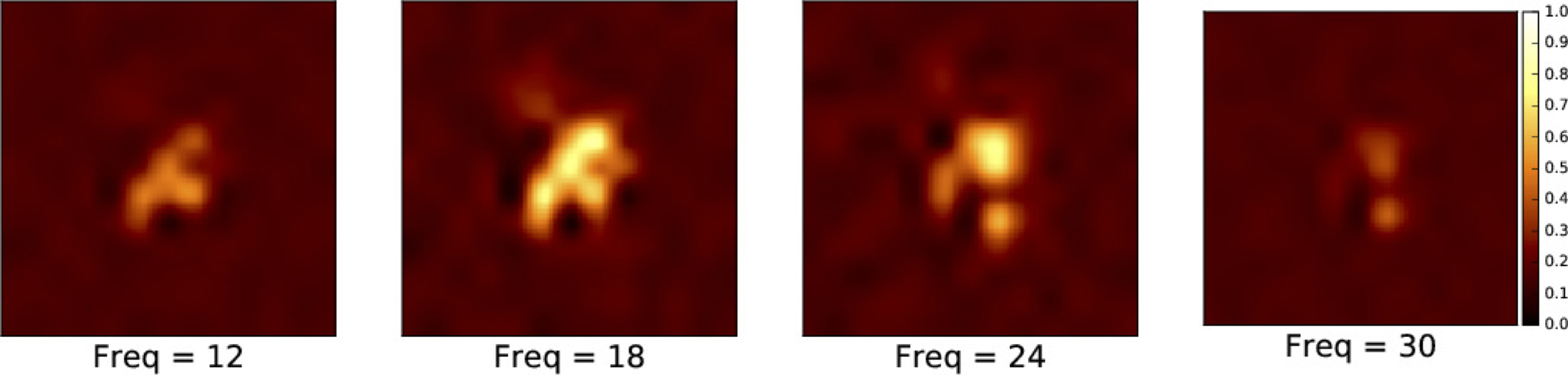}\\
   (Level 3)
    \caption{Original cuts of the data, and low-pass approximation and reconstructions for level 1 and 3.}
    \label{fig:level0}
    \end{figure}

To visualize the effects of extracting the low frequency components of the signal directly in 3D, we 
selected four slices of the cube at $\text{frequencies }= \ 12,\ 18,\ 24,\ 30$, as shown in Figure 
\ref{fig:level0}. To perform the multiresolution analysis, a 3D DWT is used with a Daubechies wavelet (\texttt{db5}). The first four levels are obtained by sequentially computing the low-pass (approximation) coefficients, with the corresponding reconstruction process. The results of that reconstructions are shown in Figure \ref{fig:level0}, for the levels $1$ and $3$.
     
As the level gets higher, less detailed data we get, i.e., the coarse-grained patterns appear. As stated in section \ref{proposal}, it is clear that in the high level representations, the clump structures are easier to identify that the ones in lower levels.

%


\subsection{The RMS at Different Levels} \label{clumping}
    Fellwalker and Clumpfind are very dependent of the RMS value of the data, in fact many of the parameters of these algorithms are set as multiples of the RMS value. Therefore, if the RMS of approximated data change with the decomposition level, then the corresponding algorithm will change its behavior.

\begin{figure}[htpb!]
    \centering
    \includegraphics[width=.84\linewidth]{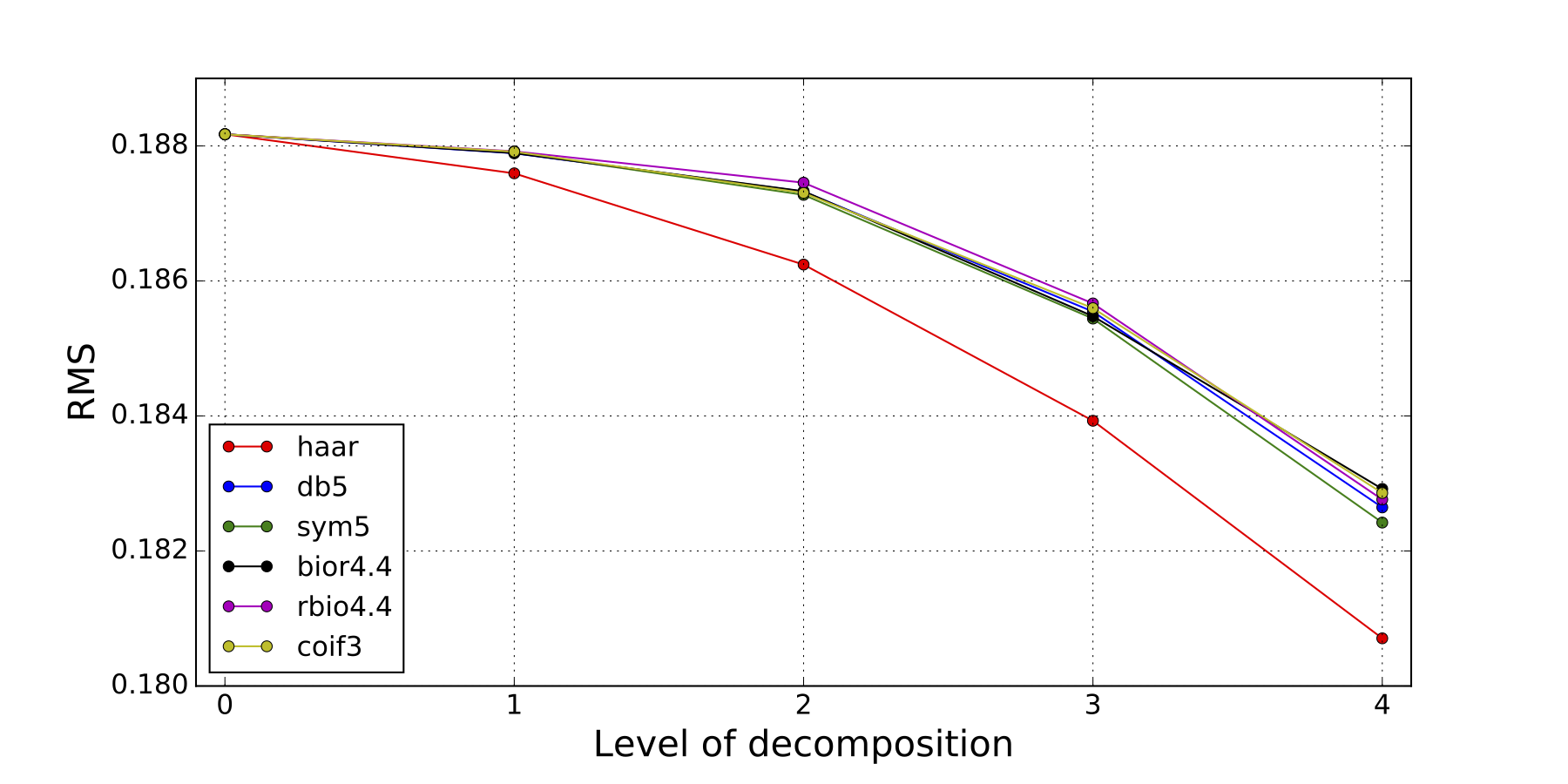}\\
    \includegraphics[width=.84\linewidth]{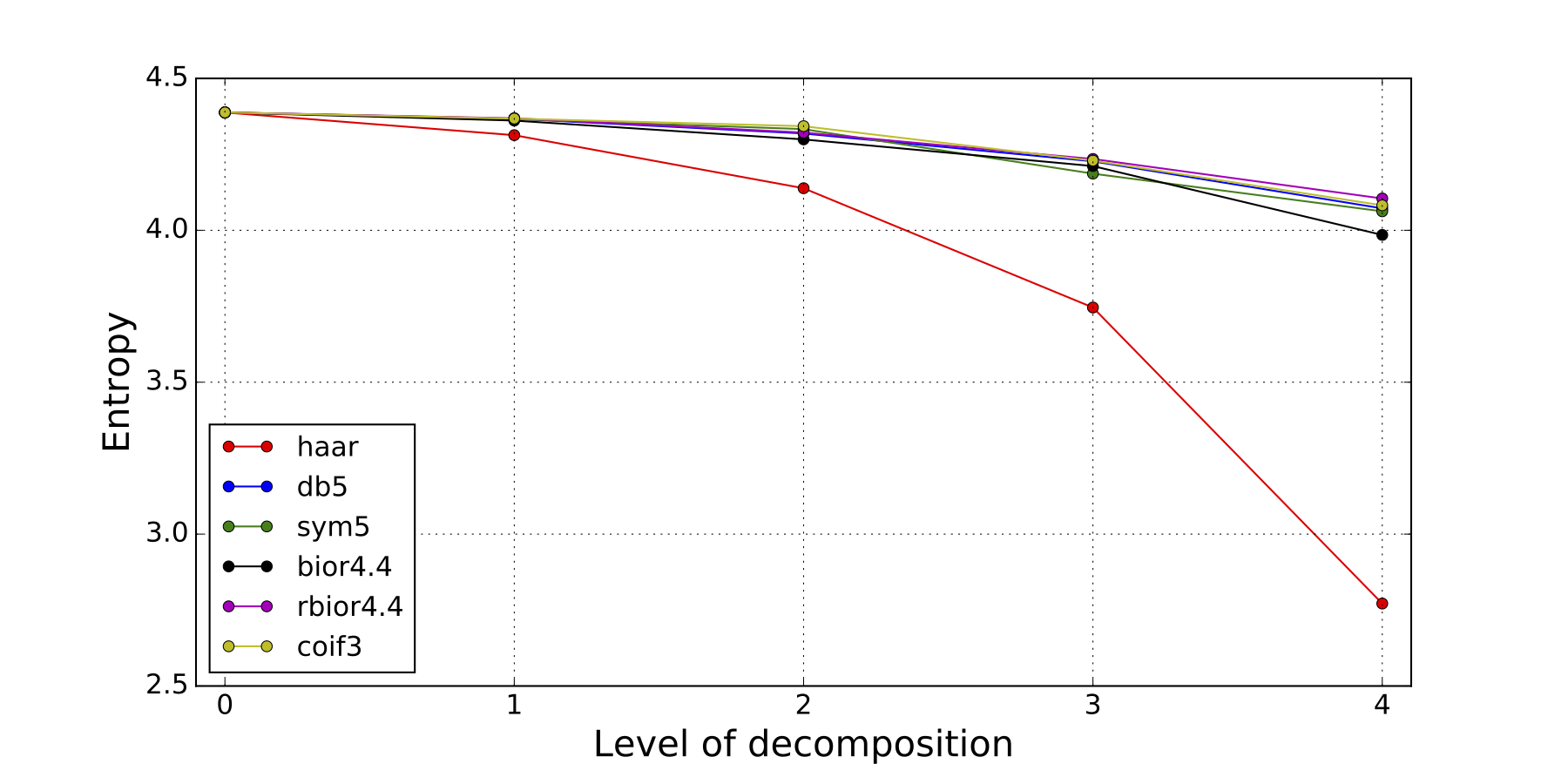}
    \caption{RMS of data decomposed at different levels (top) and Entropy (bottom) of data decomposed at different levels.}
    \label{fig:rms}
    \end{figure}

    The upper plot of Figure \ref{fig:rms} presents the RMS variation with respect to the decomposition level for each wavelet. It is clear that there is an inverse relation between them, verifying the property explained in Section \ref{SourceIdentification}. 
    
    
    Moreover, the (\emph{Shannon}) entropy also decreases with the decomposition level, as shown in the bottom part of Figure \ref{fig:rms}. This is explained because fine-grained (high frequency) details introduce more local information to the images, increasing their variability and becoming less compressible (entropic). 
    Please note the steep decay of RMS and entropy for the Haar wavelet compared with the rest. This can be explained due to the square shape of the wavelet, which does not fit the soft intensity decay on resolved astronomical images. 
    
\subsection{The Variable RMS Approach}

    To asses the contribution of computing the RMS at each level, we compare our approach to a control version of it without variable RMS, namely the \emph{fixed RMS} approach.    
    
\begin{figure}[htpb]
    \centering
    \includegraphics[width=.99\linewidth]{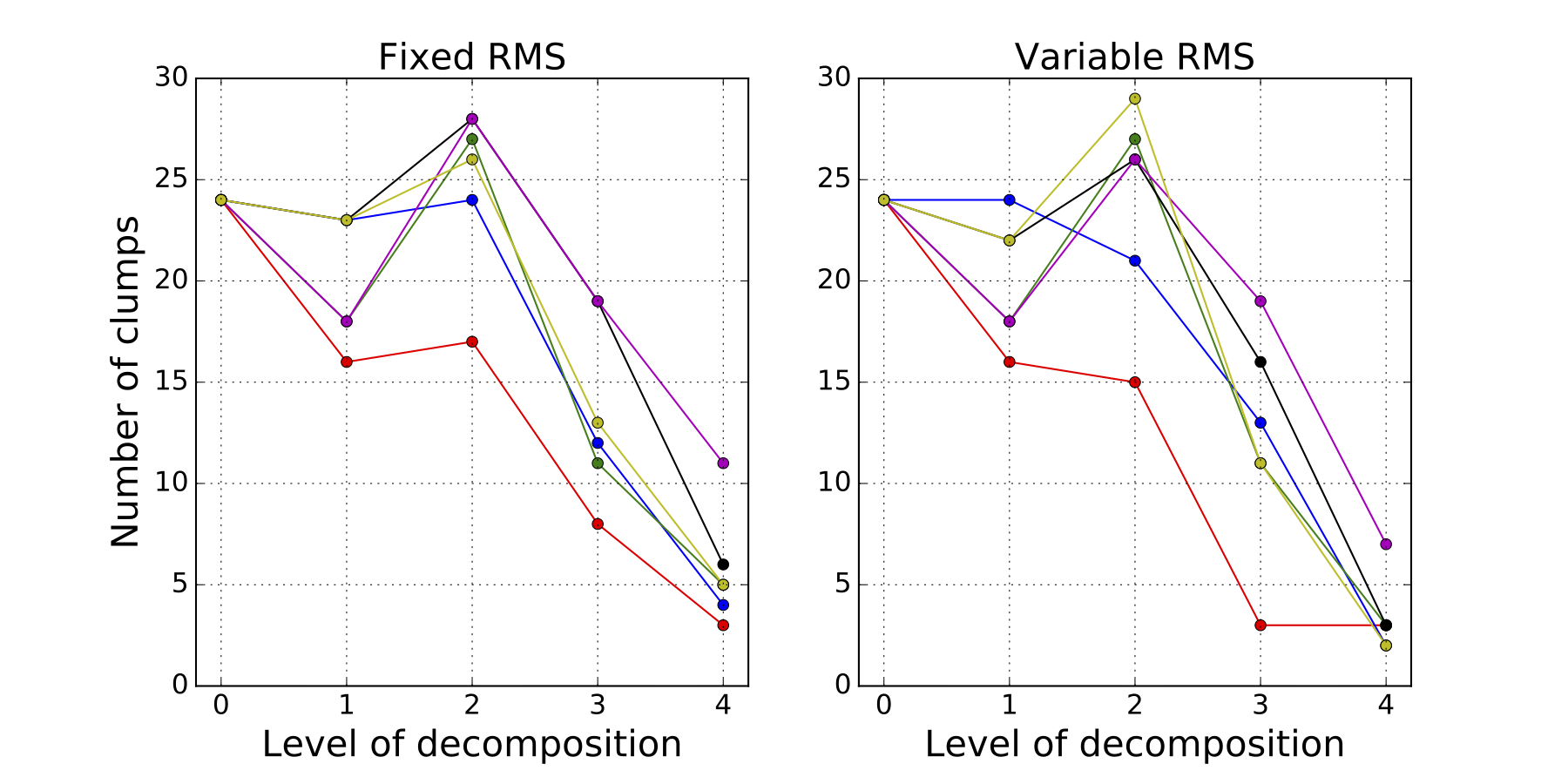}\\
    (a)\\
    \includegraphics[width=.99\linewidth]{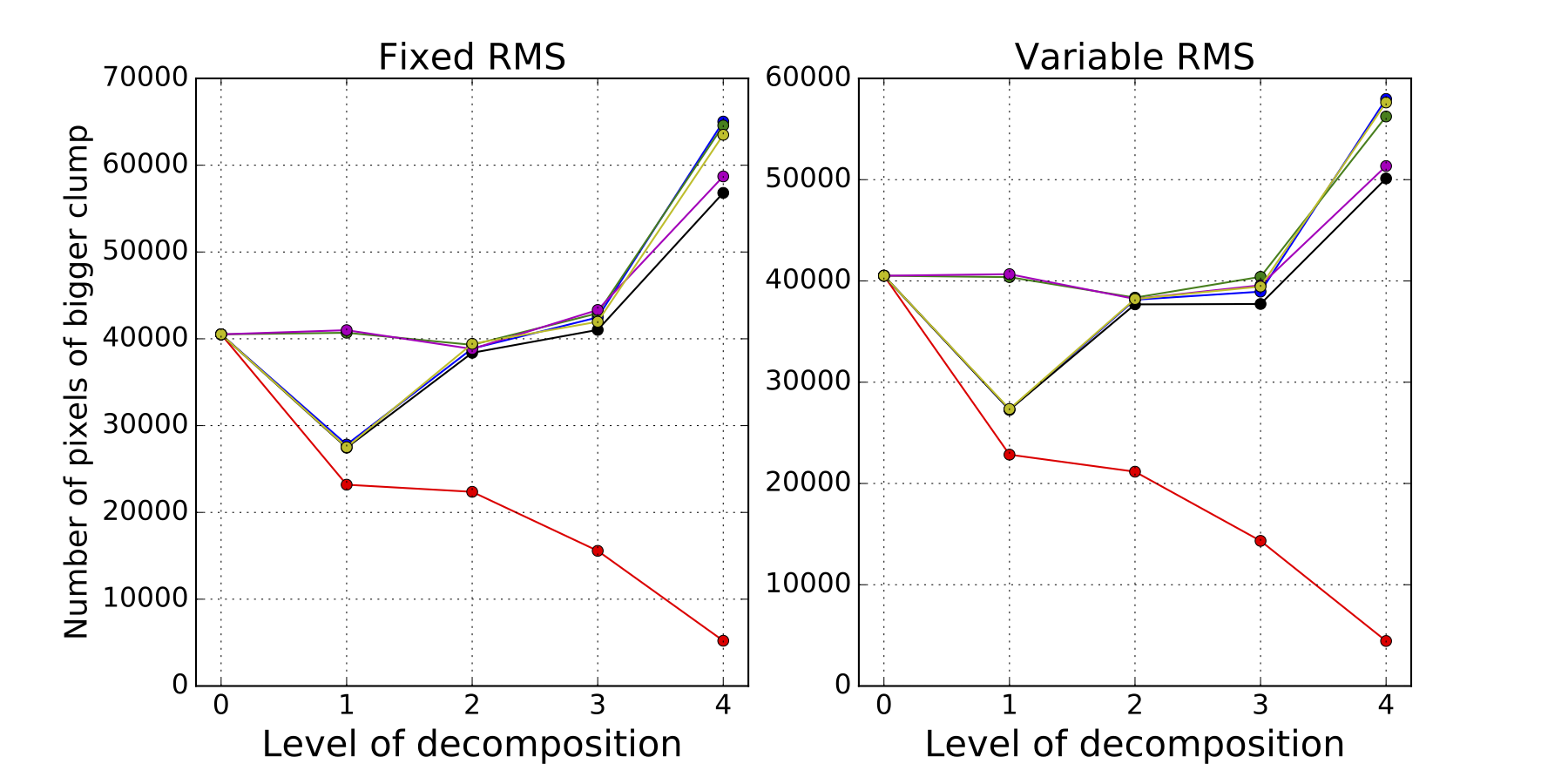}\\
    (b)\\
    \includegraphics[width=.99\linewidth]{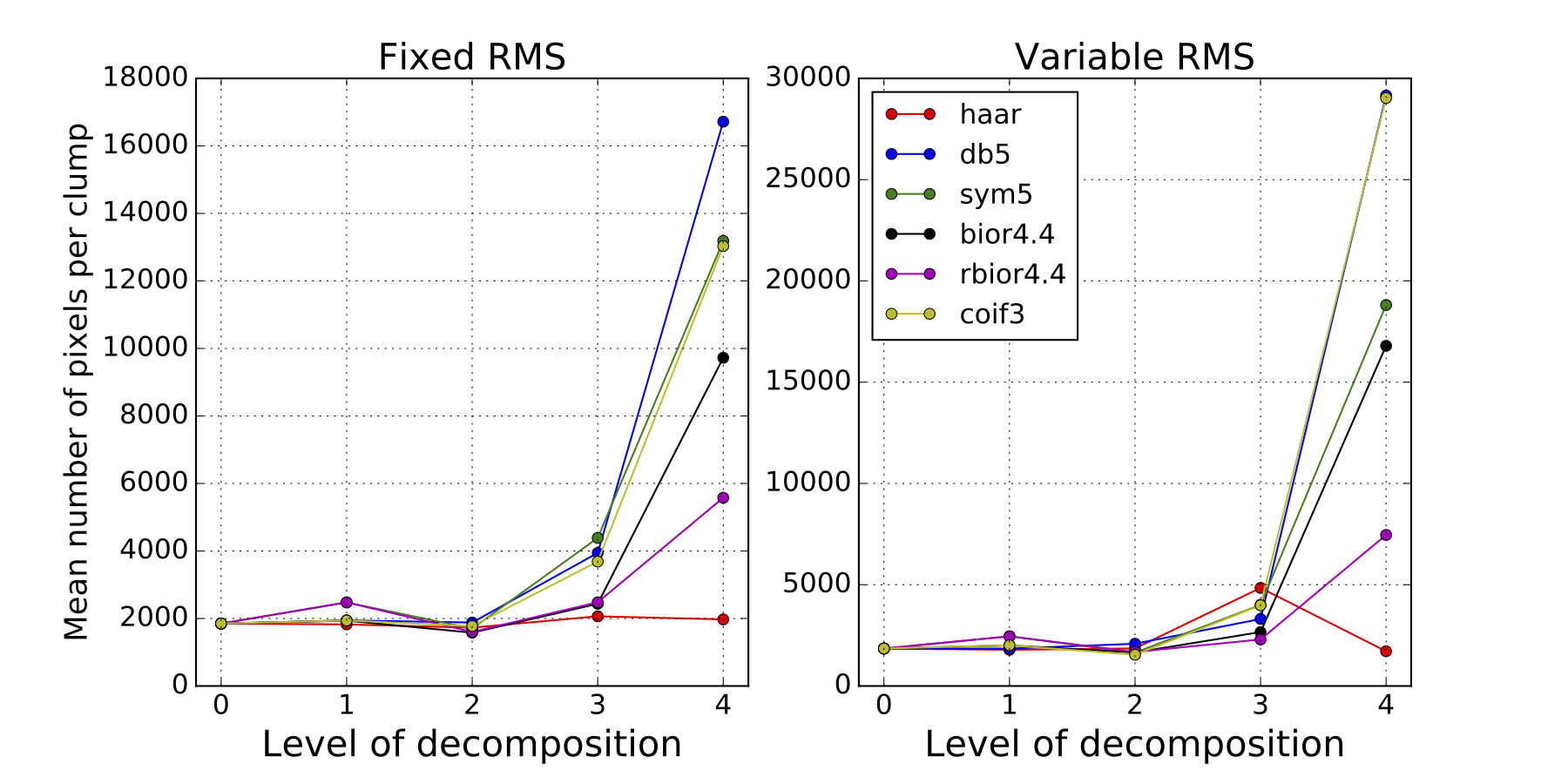}\\
    (c) 
    \caption{(a) Number of detected clumps at different levels. (b) Number of pixels of the biggest detected clump at different levels. (c) Mean number of pixels of clumps detected at different levels}
    \label{fig:meanpix}
    \end{figure}

    The obtained results are summarized in Figure \ref{fig:meanpix}, showing the number of detected clumps, number of pixels of the biggest clump and the mean number of pixels of the clumps. We can observe that there is a decreasing tendency for the number of detected clumps, which is the expected behavior. Also, there is an increasing tendency for the number of pixels of the biggest clump, and the mean number of pixels per clump, which (again) is the expected behavior. The only exception is the Haar wavelet, that shows an opposite behavior. This supports our conclusion that the Haar wavelet is not useful for processing this type of data.
        
    Noticeable differences are observed between the two approaches for the RMS value. In particular Figure \ref{fig:meanpix}(c) shows a huge change between the mean number of pixels per clump, that supports our hypothesis about the changing behavior of the clumping algorithms with the RMS. This is due to the decrease of RMS in the second approach (variable RMS) and the thresholding stage of Fellwalker, where all the pixels below some multiple of the RMS are set as unusable. 
    
    \subsection{Hierarchical Relations} 
    
\begin{figure}[htpb]
    \centering
    \includegraphics[width=.14\linewidth]{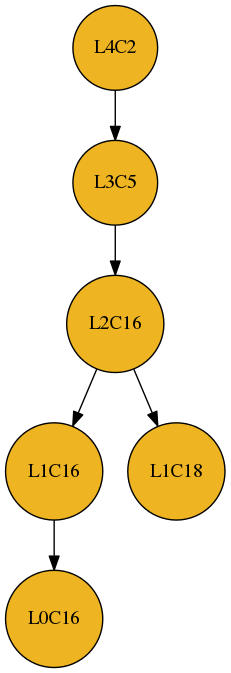}
    \includegraphics[width=.84\linewidth]{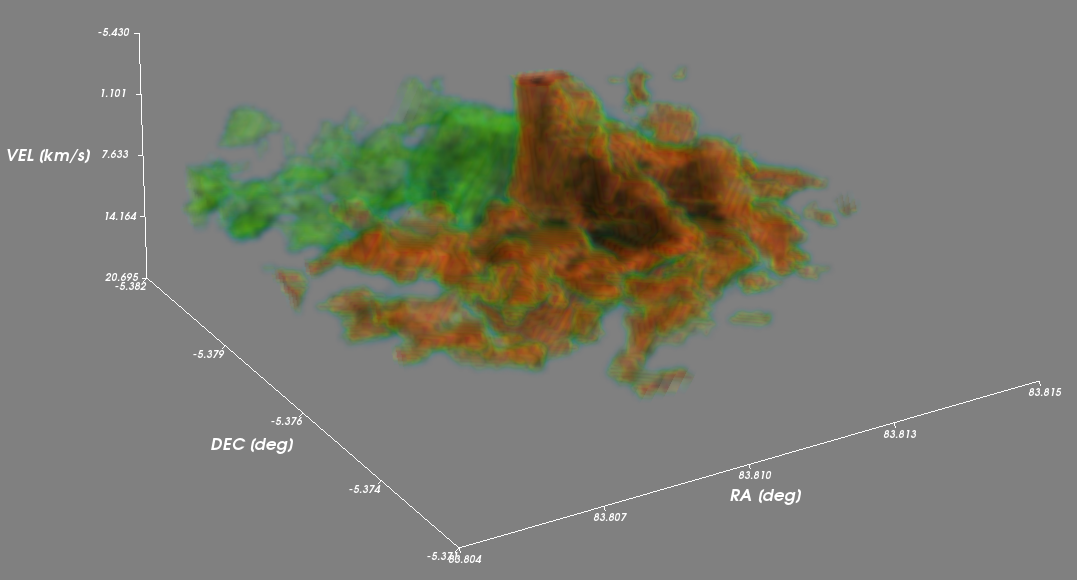}
    \caption{Portion of the hierarchical tree structure (left), and 3D visualization of L2C16, L1C16 and L1C18 (right)}
    \label{fig:3Dplot}
    \end{figure}

    For the representation of the hierarchical relations between the clumps detected at different levels, a tree structure is used as shown in Figure \ref{fig:3Dplot} (left). The clumps at the top of the three correspond to structures detected at the higher level of MRA with coarse-grained details, and the clumps at the bottom of the three are the ones detected in the original data.
    We report that in the whole hierarchical tree, there is a large number of isolated nodes/clumps that are not connected to clumps at other levels, yet this is expected as not all the detected clumps are part of the truth noiseless signal.
    
    The labels of the nodes have the following structure: $\text{L}\{\texttt{level number}\}\text{C}\{\texttt{clump identifier}\}$, indicating the level of decomposition and the identifier of the detected clump at this level.
    From the tree structure, it is possible to visualize graphically the relation between the clumps, as shown in Figure \ref{fig:3Dplot} (right). Here we plot L2C16 (cyan), L1C16 (green) and L1C18 (brown). As expected, the L2C16 clump encloses the other two, i.e, when we perform clumping with more detailed data, the original clump is splitted into two structures.


\section{Conclusions and Future Work} \label{conclusion}

In this work, we have implemented and studied a multiresolution analysis approach using different Wavelet families over 3D astronomical data. The aim was to perform clumping at different levels of details, but delivering also to the astronomer a hierarchical relation of the structures detected between levels.

The main contribution of the paper is the implementation of the multiresolution analysis over data cubes by directly using the 3D DWT, since the SWT becomes an unfeasible alternative for 3D data due to the redundancy of the approximation coefficients. We show empirically that there no much difference between choosing different Wavelet families, and only fails when choosing very sharp functions like Haar.


The strong dependency of the used clumping algorithm (Fellwalker) on the RMS was considered by dynamically setting the parameter for each level, and comparing these results to the simpler case of using a fixed RMS version of the method. 

In terms of future work, there are a few research directions that we would like to explore. We suspect that dropping all the approximation coefficients but AAA it is too drastic. It would be interesting to analyze the effect of dropping a subset of these coefficient at each level, for instance to keep \{AAA, DAA, ADA, AAD \}. In terms of clumping algorithms we have arbitrarily selected FellWalker, yet other algorithms need to be explored and compared. Also, different methods for connecting clumps at neighbouring levels could be tested, as they are only heuristic to detect the underlying relationship between clumps at different scales. At last, a more thoroughly study of the properties of the approach is needed through the use of synthetic data, because (in general) there are no ground-truth labels for astronomical data. 


\section*{Acknowledgements}
This research was possible due to CONICYT-Chile fundings, specifically through the project FONDEF IT15I10041 and the support of Basal Project FB-0821 and Basal Project FB-0008.

\begingroup
\setstretch{0.8}
\bibliographystyle{unsrt}
\bibliography{wavclumps}
\endgroup

\end{document}

%% file: preamble.tex
\usepackage[a4paper,textwidth=18cm,textheight=24cm,top=2.85cm, bottom=2.85cm, left=1.5cm, right=1.5cm]{geometry}

\usepackage{icdp2009}

\makeatletter
\long\def\@makecaption#1#2{%
\vskip\abovecaptionskip
\sbox\@tempboxa{#1. #2}%
\ifdim \wd\@tempboxa >\hsize
#1. #2\par
\else
\global \@minipagefalse
\hb@xt@\hsize{\box\@tempboxa\hfil}%
\fi
\vskip\belowcaptionskip}
\makeatother

%% file: wavclumps.bbl
\begin{thebibliography}{10}

\bibitem{Bertin}
E.~{Bertin} and S.~{Arnouts}.
\newblock {SExtractor: Software for source extraction}.
\newblock {\em {Astronomy and Astrophysics Supplement}}, 117:393--404, June
  1996.

\bibitem{Williams}
{Jonathan P. Williams, Eugene J. de Geus and Leo Blitz}.
\newblock {Determining structure in molecular clouds}.
\newblock {\em {The Astrophysical Journal}}, {498}:{693--712}, {1994}.

\bibitem{Stutzki}
{J. Stutzki and R. Güsten}.
\newblock {High spatial resolution isotopic CO and CS observations on M17 SW:
  The clumpy structure of the molecular cloud core}.
\newblock {\em {The Astrophysical Journal}}, {356}:{513--533}, {1990}.

\bibitem{Berry}
{D.S Berry}.
\newblock {FellWalker—A clump identification algorithm}.
\newblock {\em {Astronomy and Computing}}, {10}:{22--31}, {2015}.

\bibitem{Rosolowsky}
{E.W Rosolowsky, J.E Pineda, J. Kauffmann and A. A. Goodman}.
\newblock {Structural Analysis of Molecular Clouds: Dendrograms}.
\newblock {\em {The Astrophysical Journal}}, {679}:{1338--1351}, {2008}.

\bibitem{Starck}
{J.-L Starck and F. Murtagh}.
\newblock {\em {Astronomical Image and Data Analysis}}.
\newblock {Springer-Verlag}, {2nd edition} edition, {2006}.

\bibitem{Kennicutt}
{R. Kennicutt Jr and N. Evans II}.
\newblock {Star Formation in the Milky Way and Nearby Galaxies}.
\newblock {\em {Anual Review of Astronomy and Astrophysics}}, {50}:{531--608},
  {2012}.

\bibitem{Alves}
{J. Alves, M. Lombardi and C. Lada}.
\newblock {The mass function of dense molecular cores and the origin of the
  IMF}.
\newblock {\em {Astronomy and Astrophysics}}, {462}:{17--21}, {2007}.

\bibitem{Gregorio}
{ Rodrigo Gregorio, Mauricio Solar, Diego Mardones, Karim Pichara, Victor
  Parada and Ricardo Contreras }.
\newblock {Automatic detection and automatic classification of structures in
  astronomical images}, {2014}.

\bibitem{Daubechies}
{Daubechies, I.}
\newblock {\em {Ten Lectures on Wavelets}}.
\newblock {Society for Industrial and Applied Mathematics}, 1992.

\end{thebibliography}
